\documentclass[preprint,12pt]{elsarticle}

\journal{Information Fusion}

\makeatletter
\def\ps@pprintTitle{%
  \let\@oddhead\@empty
  \let\@evenhead\@empty
  \let\@oddfoot\@empty
  \let\@evenfoot\@empty
}
\makeatother

\usepackage{amsmath,amssymb,amsfonts,mathtools}
\usepackage{graphicx}
\usepackage{booktabs}
\usepackage{multirow}
\usepackage[table]{xcolor}
\usepackage{array}
\usepackage{makecell}
\usepackage{siunitx}
\usepackage{float}
\usepackage{textcomp}
\usepackage[T1]{fontenc}

\renewcommand{\vec}[1]{\boldsymbol{#1}}

\newcommand{\TABabbr}{\mbox{\emph{TAB\_}\/}}
\newcolumntype{C}[1]{>{\centering\arraybackslash}p{#1}}

\graphicspath{{figures/}}

\begin{document}

\begin{frontmatter}

\title{Label-Conditioned Cross-Modal Fusion for Adult-to-Pediatric ECG Transfer via Curriculum-Gated Contrastive Alignment}

\author[seu]{Xinran Liu\fnref{fn1}}
\author[seu]{Yuwen Li\fnref{fn1}}
\author[seu]{Hongxiang Gao}
\author[seu]{Heyang Xu}
\author[nmu]{Jianqing Li}
\author[zzu]{Zongmin Wang\corref{cor2}}
\author[seu]{Chengyu Liu\corref{cor1}}

\cortext[cor1]{Corresponding author. Email: chengyu@seu.edu.cn (Chengyu Liu)}
\cortext[cor2]{Corresponding author. Email: zmwang@zzu.edu.cn (Zongmin Wang)}

\affiliation[seu]{organization={School of Instrument Science and Engineering, Southeast University},
            addressline={Nanjing 210096},
            city={Nanjing},
            country={China}}

\affiliation[nmu]{organization={Nanjing Medical University},
            addressline={Nanjing 211166},
            city={Nanjing},
            country={China}}

\affiliation[zzu]{organization={Zhengzhou University},
            addressline={Zhengzhou 450001},
            city={Zhengzhou},
            country={China}}

\fntext[fn1]{Xinran Liu and Yuwen Li contributed equally to this work.}

\begin{abstract}
Automated pediatric electrocardiogram (ECG) interpretation remains challenging because developmental differences in heart rate, intervals, and waveforms limit the transferability of models trained mainly on adult data, while expert-labeled pediatric ECG cohorts are scarce. We propose \textbf{PEACE} (\textbf{P}ediatric-\textbf{A}dult ECG \textbf{A}lignment via \textbf{C}ross-modal \textbf{E}nhancement), an adult-to-pediatric ECG transfer framework pretrained on MIMIC-IV ECGs and adapted to pediatric targets. PEACE integrates label-specific bidirectional contrastive learning (LSBC) to align ECG representations with diagnostic semantics and curriculum adaptive fusion (CAF) to stabilize optimization under limited pediatric supervision. Label-conditioned short text descriptors provide auxiliary semantic supervision during training, whereas inference requires ECG signals only. On ZZU-pECG, PEACE achieves macro-average AUCs of 59.39\%, 81.74\%, and 91.56\% under zero-shot, 50-shot, and full fine-tuning settings, respectively, outperforming ECG-only, multimodal, and generic domain adaptation baselines including DANN and MMD. On PTB-XL, it reaches 96.90\% macro-average AUC after full fine-tuning over nine harmonized labels with nonzero mapped incidence. Gradient-based attention maps show increased saliency around QRS voltage and morphology regions for chamber-related \emph{RVH} and around QRS-to-T/repolarization intervals for \emph{LQTS}, broadly consistent with ECG regions commonly inspected during routine interpretation. These results suggest that adult-scale ECG pretraining coupled with rhythm, morphology, and ST-T repolarization semantic descriptors improves transferable pediatric diagnosis under label scarcity while preserving clinically interpretable waveform focus.
\end{abstract}


\end{frontmatter}

\section{Introduction}

Transfer across populations is challenging when the source domain is large and semantically annotated, while the target domain exhibits distribution shift and labels are scarce. We study pediatric ECG classification, where adult ECG datasets are abundant but pediatric labels are limited and few cohorts provide paired free-text interpretations for every tracing. Distribution mismatch induced by developmental electrophysiology causes adult-trained models to degrade on pediatric signals \cite{ribeiro2020automatic, children12010025, chen2024congenital}, and limited pediatric supervision further constrains adaptation \cite{ZHU202287}.
Clinically, pediatric ECG interpretation differs from adult interpretation because normal rate, intervals, voltage criteria, conduction patterns, and repolarization morphology vary with development. As a result, models trained primarily on adult cohorts may transfer poorly when pediatric labels are scarce. Methodologically, this defines a supervised transfer setting in which large adult ECG resources provide useful priors, but the pediatric target requires adaptation that respects age-dependent waveform morphology and label imbalance.

Recent high-capacity models for arrhythmia detection in adults \cite{chen2023srecg, children12010025, ECGLM_Sci2024} perform well in-domain, yet their efficacy diminishes in pediatric scenarios when morphology lies outside the training distribution \cite{children12010025}. While recent efforts have explored generalization across racial groups \cite{2024KED}, transfer from adult to pediatric cohorts under severe distribution shift remains less studied \cite{liang2021domain}.
To address this challenge, we propose \textbf{PEACE}, \textbf{Pediatric-Adult ECG Alignment via Cross-modal Enhancement}, a structured multimodal framework for adult to pediatric ECG transfer. PEACE uses label conditioned semantic descriptors as supervision used only during training to anchor ECG representations to clinically meaningful rhythm, morphology, and ST-T repolarization evidence, while inference remains ECG only. By introducing structured semantic supervision and alignment guided by labels, PEACE learns representations that are more robust to pediatric distribution shift.

Beyond aggregate macro average scores, PEACE shows higher per-class AUC than the evaluated baselines on several pediatric categories, notably labels related to conduction and repolarization, such as \emph{LAFB}, \emph{IRBBB}, \emph{LQTS}, and \emph{STTC}, while remaining competitive on high prevalence classes such as \TABabbr. Per-label AUC across zero-shot, few-shot, and full fine-tuning is reported in Table~\ref{tab:per_class}.

The technical core of PEACE comprises two components: Label Specific Bidirectional Contrastive Learning (LSBC) and Curriculum Adaptive Fusion (CAF). LSBC constructs disease aware positive pairs using three complementary semantic descriptor streams per label: rhythm, morphology, and ST-T repolarization, aligning waveform features with structured text supervision rather than a single fused prompt. CAF employs a multi stage curriculum to dynamically calibrate modality weights. This process stabilizes transfer across populations and mitigates representation disruption during early training.

\begin{figure}[!t]
\centering
\includegraphics[width=\linewidth]{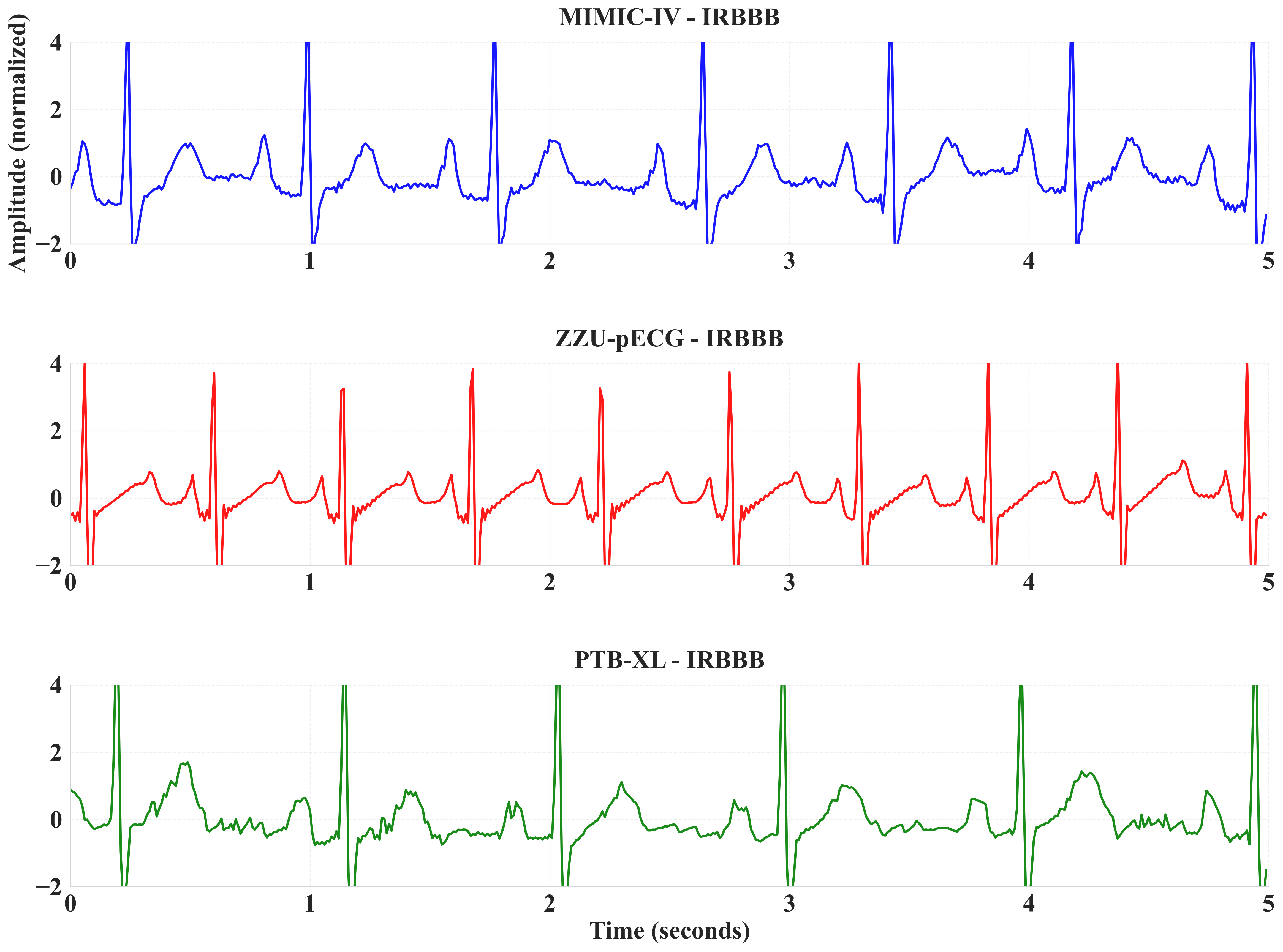}
\caption{\emph{IRBBB} morphology on Lead II (five-second segments) across populations. Top: MIMIC-IV adult; middle: ZZU-pECG pediatric; bottom: PTB-XL adult.}
\label{fig:irbbb_shift}
\end{figure}

Figure~\ref{fig:irbbb_shift} contrasts five-second Lead II segments of \emph{IRBBB} across adult MIMIC-IV, pediatric ZZU-pECG, and adult PTB-XL. The pediatric trace shows faster heart rate and interval scaling relative to adult traces for the same diagnostic label, illustrating that transfer across populations must accommodate joint variation in rate and morphology rather than label identity alone.

We validate PEACE on ZZU-pECG \cite{2025zzu_pecg}, a high-fidelity pediatric database comprising ECG records from 11,643 children. We additionally evaluate on PTB-XL \cite{Wagner2020PTBXL}; external macro average AUC is reported over the nine harmonized labels with nonzero mapped incidence under our SCP to ontology mapping (Sec.~\ref{experiment_dataset}). PEACE achieves higher macro average AUC than the evaluated baselines on these benchmarks.

Taken together, this study examines whether label-conditioned cross-modal fusion can make adult-scale ECG pretraining transferable to pediatric ECG interpretation beyond what is achievable by generic domain adaptation, global multimodal alignment, or naive ECG--text fusion. PEACE is not a conventional ECG-plus-text input fusion system. Instead, it uses label-specific semantic descriptors as a privileged training modality: diagnostic labels act as fusion queries, ECG and semantic representations are aligned at the label level, and curriculum gating regulates the strength of cross-modal alignment during optimization. At inference, the semantic branch is discarded and predictions are made from ECG signals alone. This design isolates the benefit of training-time label-conditioned fusion without introducing additional deployment-time modality requirements.

The primary contributions are summarized as follows:
\begin{itemize}
\item \textbf{A label-conditioned multimodal framework for adult-to-pediatric ECG transfer.} We formulate pediatric ECG diagnosis as a cross-population transfer problem in which adult-scale ECG pretraining is guided by label-conditioned semantic supervision during training, while inference remains ECG-only. This design leverages clinical semantics without introducing text dependence at deployment.

\item \textbf{Three-axis semantic descriptors for structured clinical evidence modeling.} For each diagnostic label, we construct rhythm, morphology, and ST-T repolarization descriptors that reflect complementary steps of ECG interpretation. Compared with a single fused prompt, this decomposition provides a more structured semantic space and reduces evidence entanglement in multi-label ECG classification.

\item \textbf{Label-query alignment with curriculum-adaptive optimization.} We design a Label-Query Network (LQN) to extract label-specific ECG and semantic features, and introduce Label-Specific Bidirectional Contrastive Learning (LSBC) to align waveform and semantic representations at the label level. Curriculum Adaptive Fusion (CAF) further regulates the alignment strength during training, reducing disruption of pretrained ECG features under limited pediatric supervision.

\item \textbf{Comprehensive evaluation against transfer, fusion, and ECG pretraining baselines.} Beyond ECG foundation-model and semantic pretraining baselines, we compare PEACE with canonical domain-adaptation methods and standard early/late fusion strategies under the same ResNet1D initialization and ZZU-pECG protocol. This evaluation isolates the benefit of PEACE's label-conditioned semantic alignment mechanism.
\end{itemize}

\begin{figure}[!t]
\centering
  \includegraphics[width=\textwidth]{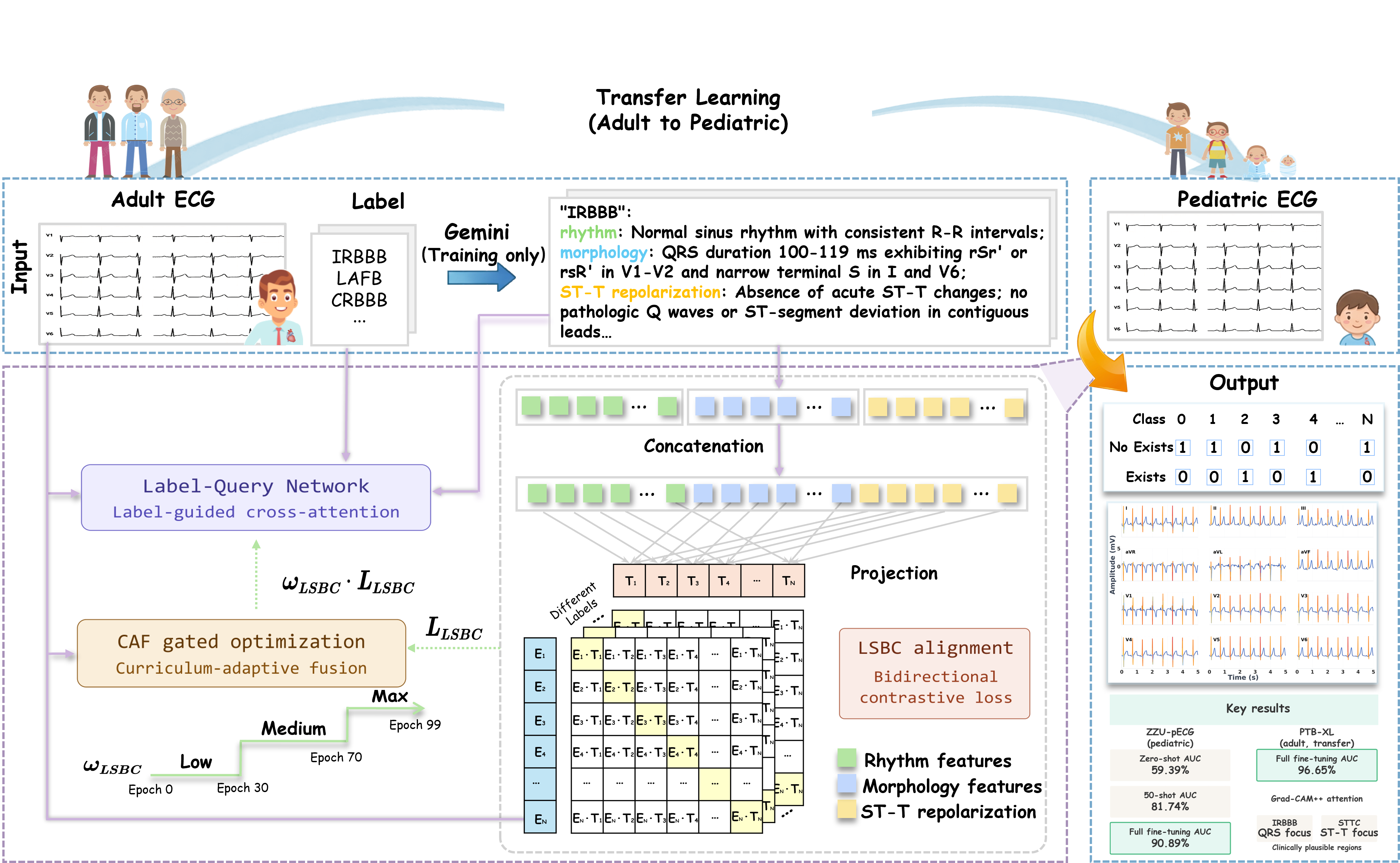}
\caption{Overview of PEACE. Label conditioned semantic descriptors are used only during training to guide cross-modal alignment; at inference, the semantic branch is discarded and predictions are made from ECG signals alone.}
\label{FIG:PEACE}
\end{figure}

\section{Related Work}
\subsection{ECG Representation Learning}
ECG analysis evolved from foundational CNN and RNN architectures \cite{Kiranyaz, YILDIRIM, CHEN} to attention-based Transformer paradigms \cite{transformer, 2024ECGformer}. Masked reconstruction and related self-supervised objectives have also been explored for ECG waveforms, including masked auto-encoding frameworks \cite{STMEM}. These methods capture temporal dependencies, but direct transfer from corpora dominated by adults to pediatric cohorts remains difficult under developmental distribution shift.

\subsection{Multimodal ECG and text alignment}
Early integration relied on coarse strategies such as feature concatenation or late fusion \cite{lalam2023ecg}. These often fail to capture interactions between waveforms and clinical text. Contemporary frameworks \cite{2024KED, Li2025ECGFounder, wang2022medclip} use attention-based global alignment, but pediatric transfer requires finer correspondence between local waveform structure and semantics because morphology deviates from adult-centered patterns \cite{gliner2025clinically}.

\subsection{Label Semantics and Label-Aware Contrastive Learning}

Contrastive learning has been widely adopted to improve class-discriminative representations by pulling semantically related samples closer while separating unrelated samples. In medical and multi-label learning scenarios, prior studies have incorporated supervised diagnostic hierarchies \cite{9874526, NEURIPS2023_d74f9efa}, multi-task diagnostic objectives \cite{hou2023mtdiag}, and bidirectional or group-wise alignment strategies \cite{BiG}. These methods show that label information can provide useful structure for representation learning. However, most existing formulations still operate on global sample-level embeddings or use labels mainly to define positive and negative pairs, which may underutilize label semantics when target-domain supervision is scarce.

This limitation is particularly relevant to adult-to-pediatric ECG transfer. Pediatric ECGs differ from adult ECGs not only in global data distribution, but also in the physiological meaning of rate, intervals, voltage, conduction morphology, and repolarization patterns. A global contrastive objective may align samples at the representation level, but it does not explicitly indicate which waveform evidence should support each diagnostic label. Similarly, using a single textual descriptor per class may collapse rhythm, morphology, and ST-T repolarization evidence into an undifferentiated semantic vector, increasing evidence entanglement in multi-label ECG classification.

Our LSBC objective follows supervised contrastive learning in spirit \cite{khosla2020supervised}, but adapts it to label-conditioned ECG transfer in three aspects. First, alignment is performed on queried features conditioned by diagnostic labels rather than on global pooled representations. Second, semantic supervision uses three axis-specific descriptors per label, corresponding to rhythm, morphology, and ST-T repolarization, rather than one fused text stream. Third, the alignment intensity is dynamically gated by CAF to reduce disruption of pretrained ECG features during early adaptation. Therefore, the contribution is not a standalone generic contrastive primitive, but a structured label-conditioned alignment strategy that integrates LQN-based feature probing, evidence-axis semantic descriptors, LSBC, and curriculum-gated optimization for low-resource pediatric ECG transfer.

\subsection{Transfer across populations under target scarcity}
Recent advances explore spatiotemporal memory \cite{STMEM}, self-supervised pretraining \cite{MERL}, structured knowledge injection \cite{tang2026interpretable}, medical prompting \cite{wang2025ecg}, and language supervision \cite{zhou2025diagnosis}. Curriculum-based denoising also improves multimodal optimization \cite{xu2025multimodal}. Our setting differs in that the target domain is pediatric, strongly shifted, and has few labels; this motivates alignment conditioned on labels plus curriculum gating rather than relying on global alignment alone.

\subsection{Domain adaptation perspective}
Physiological discrepancies across heart rate, morphology, and thresholds create substantial distribution shift between children and adults \cite{children12010025, Heart2005}. Developmental shifts, such as neonatal ventricular dominance, require age-aware interpretation criteria. Models trained predominantly on adult data therefore show notable mismatch on pediatric cohorts \cite{Gow2023MIMICIVECG, PMID:37879933}.

Existing adaptation techniques target adult scenarios via unsupervised alignment \cite{HE2023119711}, patient reweighting \cite{FAN2025126824}, or distribution matching \cite{long2015learning}. Standard diagnostic guidelines \cite{surawicz2009aha} assume relatively static physiology, which conventional strategies struggle to reconcile with pediatric maturation. These methods are ill equipped for pediatric transfer, where pathological manifestations are governed by maturational mechanisms and symptomatic overlaps with conditions like Brugada syndrome \cite{brugada1992right}.

PEACE performs supervised transfer across populations. LSBC uses feature matching specific to each label to mitigate pediatric shift, and CAF modulates alignment strength to improve stability during adaptation from adult to pediatric data \cite{xu2025multimodal, liu2025knowledge}.

\section{Methodology}
\subsection{Overview}
The PEACE framework targets age related waveform morphology shift, i.e., the physiological variation in ECG rate, intervals, and morphology between adult and pediatric populations. As illustrated in Figure~\ref{FIG:PEACE}, adult ECGs serve as the source of physiological knowledge, while diagnostic labels provide dual purpose supervision for multi-label classification and the derivation of semantic prompts driven by labels. PEACE mimics the clinical diagnostic process by isolating pathological markers through the LQN, aligning them via LSBC, and stabilizing the transfer via CAF.

\textbf{Notation.} We denote batch size by $B$, the number of ECG time tokens by $T$, and the number of diagnostic labels by $C$. For mathematical simplicity, we set the latent dimensions of all encoders to be equal, i.e., $d_{\mathrm{ecg}} = d_{\mathrm{lbl}} = d$ for per-token or per-axis embeddings, and $d_{\mathrm{share}}$ after projection. The ECG encoder output is denoted by $\vec{X}_{\mathrm{ecg}} \in \mathbb{R}^{B \times T \times d}$, the semantic descriptor sequence by $\vec{X}_{\mathrm{rep}} \in \mathbb{R}^{B \times 3 \times d}$, with one token for each of the rhythm, morphology, and ST-T repolarization descriptor streams, and label text embeddings by $\vec{Z}_{\mathrm{lbl}} \in \mathbb{R}^{C \times d}$. We keep the symbol ``rep'' for continuity with earlier drafts; this branch encodes Gemini-derived label-level descriptors, not patient-specific clinical reports. Subscripts $\mathrm{morph}$ and $\mathrm{STT}$ abbreviate the morphology and ST-T repolarization axes, respectively; $\mathrm{rhythm}$ is left unabbreviated. Projected tokens use dimension $d_{\mathrm{share}}$ with $d_{\mathrm{share}} \ll d$ and $d_{\mathrm{share}} \in \mathbb{N}^+$. We use $\mathrm{sim}(\cdot,\cdot;\tau)$ for cosine similarity with temperature $\tau$.

\subsection{Multimodal encoders}
\textbf{ECG Encoder.} We employ the \texttt{resnet1d\_wang} implementation in \texttt{torch\_ecg}, based on a 1D ResNet architecture \cite{resnetwang}, to project raw 12-lead ECG signals into token-level representations $\vec{X}_{\mathrm{ecg}} \in \mathbb{R}^{B \times T \times d}$. Each token represents a localized physiological segment and captures both fine-scale morphologies such as QRS spikes and broader temporal dynamics. Global average pooling yields the classification embedding
\vspace{-8pt}
\begin{equation}
\bar{\vec{X}}_{\mathrm{ecg}} = \frac{1}{T} \sum_{t=1}^{T} \vec{X}_{\mathrm{ecg}}[:,t,:] \in \mathbb{R}^{B \times d}.
\end{equation}
while the full token sequence $\vec{X}_{\mathrm{ecg}} \in \mathbb{R}^{B \times T \times d}$ is preserved for LQN cross-attention with label queries.
\textbf{Semantic Descriptor Encoder.} Since ZZU-pECG does not contain paired free text reports, we generate label conditioned semantic descriptors using Gemini. For each diagnostic label, Gemini produces three short descriptors along rhythm, morphology, and ST-T repolarization axes. These axes are not mutually exclusive diagnostic categories; rather, they correspond to complementary steps in ECG interpretation. For example, a conduction block may mainly depend on QRS morphology but can still be described in terms of rhythm stability and secondary ST-T changes, while long QT syndrome is driven chiefly by repolarization and interval timing but also involves temporal organization of the cardiac cycle.

The three descriptors are encoded separately using BioClinicalBERT \cite{2019bert}:
\begin{equation}
\vec{X}_{\mathrm{rhythm}}, \vec{X}_{\mathrm{morph}}, \vec{X}_{\mathrm{STT}} \in \mathbb{R}^{B \times d}.
\end{equation}
We then stack them as a three-token semantic sequence:
\begin{equation}
\vec{X}_{\mathrm{rep}} =
\mathrm{stack}(\vec{X}_{\mathrm{rhythm}}, \vec{X}_{\mathrm{morph}}, \vec{X}_{\mathrm{STT}})
\in \mathbb{R}^{B \times 3 \times d}.
\end{equation}
This representation allows the label query network to attend to different clinical evidence axes without forcing each diagnosis into a single exclusive category. Here $\vec{X}_{\mathrm{STT}}$ denotes the ST-T repolarization axis and targets wording about ST segments, T and U waves, and QT or QTc intervals; it is not treated as a proxy for adjudicated ischemia.

\textbf{Label embeddings and shared projection.} Each diagnostic label is encoded via BioClinicalBERT to produce $\vec{Z}_{\mathrm{lbl}} \in \mathbb{R}^{C \times d}$. Task specific MLP heads map modalities into $\mathbb{R}^{d_{\mathrm{share}}}$: $\tilde{\vec{X}}_{\mathrm{ecg}} = \text{MLP}_{\mathrm{ecg}}(\vec{X}_{\mathrm{ecg}})$ with pointwise application along the time dimension; $\tilde{\vec{X}}_{\mathrm{rep}} = \text{MLP}_{\mathrm{rep}}(\vec{X}_{\mathrm{rep}})$ applied independently to each of the three axis tokens, yielding $\tilde{\vec{X}}_{\mathrm{rep}} \in \mathbb{R}^{B \times 3 \times d_{\mathrm{share}}}$; and $\tilde{\vec{Z}}_{\mathrm{lbl}} = \text{MLP}_{\mathrm{lbl}}(\vec{Z}_{\mathrm{lbl}})$. The semantic branch classification loss pools axes by averaging,
\begin{equation}
\bar{\vec{X}}_{\mathrm{rep}} = \frac{1}{3} \sum_{k=1}^{3} \tilde{\vec{X}}_{\mathrm{rep}}[:,k,:] \in \mathbb{R}^{B \times d_{\mathrm{share}}},
\end{equation}
before the multi-label head. For the LQN, keys and values are built from the full time sequence $\tilde{\vec{X}}_{\mathrm{ecg}} \in \mathbb{R}^{B \times T \times d_{\mathrm{share}}}$ and from the three semantic tokens $\tilde{\vec{X}}_{\mathrm{rep}} \in \mathbb{R}^{B \times 3 \times d_{\mathrm{share}}}$.

\textbf{Gemini prompting.} Axis specific strings are produced under clinical reporting constraints using a shared instruction stem with axis focused follow ups below. These strings provide privileged semantic supervision during training only; inference remains ECG only. No-text, single-text with one fused descriptor per label, and full three axis variants are compared in Table~\ref{tab:ablation}.

For each label and each axis, we query Gemini with a shared instruction stem in which the placeholder \texttt{<LABEL>} is replaced by the target class name, followed by brief follow ups tailored to each axis: rate and rhythm intervals for the rhythm axis; QRS morphology and conduction patterns for the morphology axis; and ST-segment, T-wave, and QT wording for the ST-T repolarization axis. The stem is reproduced below in a framed monospace block.

\begin{center}
\setlength{\fboxsep}{5pt}%
\setlength{\fboxrule}{0.4pt}%
\fbox{%
\begin{minipage}{0.94\linewidth}
\footnotesize\ttfamily\raggedright\setlength{\parindent}{0pt}%
I want you to play the role of a professional electrocardiologist,
and I need you to teach me how to diagnose \texttt{<LABEL>} from 12-lead ECG,
such as what leads or what features to focus on, etc.
Your answer must be less than 50 words.
\end{minipage}%
}%
\end{center}

Representative templates follow the rhythm, morphology, and ST-T repolarization layout exemplified below: \emph{IRBBB}: sinus rhythm; QRS 100 to 119\,ms with rSr$'$ in V1 and V2; no acute ST-T deviation. \emph{CRBBB}: regular sinus rhythm; QRS $\geq$120\,ms with rsR$'$ in V1 and V2 and wide terminal S in lateral leads; secondary T-wave changes in V1 to V3 without territorial ST elevation. \emph{NORM}: sinus rhythm within conventional text ranges used at training time; normal P, PR, and QRS intervals and axis; no pathologic Q waves or diagnostic ST-T abnormalities. Third-axis prompts stress ST segments, T waves, and QT related intervals; model outputs are interpreted as descriptive text supervision rather than adjudicated ischemia. The complete label-specific semantic descriptors used in this study are provided in Appendix~A, and the raw string-to-label harmonization maps will be released with the code for reproducibility.

\subsection{Label-Query Network (LQN): task-steered feature probing}
Standard representations often suffer from \textbf{feature entanglement} in multi-label scenarios, where markers for different pathologies such as tachycardia and hypertrophy overlap in a single vector. The LQN resolves this by treating each diagnostic label as a query that extracts features specific to that label from modality embeddings via cross-attention.

For each diagnostic label $c \in \{1,2,\dots,C\}$ and sample $i \in \{1,2,\dots,B\}$, we leverage the projected label embedding $\tilde{\vec{Z}}_{\mathrm{lbl}}[c,:]$ as a query to interrogate the modality keys and values:
\vspace{-6pt}
\begin{equation}
\begin{split}
\bar{\vec{z}}_{i,c}^{\mathrm{ecg}} &= \mathrm{LN} \left( \mathrm{Attn} \left( \tilde{\vec{Z}}_{\mathrm{lbl}}[c,:], \tilde{\vec{X}}_{\mathrm{ecg}}[i,:] \right) \right) \in \mathbb{R}^{d_{\mathrm{share}}}, \\
\bar{\vec{z}}_{i,c}^{\mathrm{rep}} &= \mathrm{LN} \left( \mathrm{Attn} \left( \tilde{\vec{Z}}_{\mathrm{lbl}}[c,:], \tilde{\vec{X}}_{\mathrm{rep}}[i,:] \right) \right) \in \mathbb{R}^{d_{\mathrm{share}}}.
\end{split}
\end{equation}
where $\mathrm{Attn}(\cdot,\cdot)$ denotes the Multi-Head Cross-Attention operator with layer normalization $\mathrm{LN}(\cdot)$, and $\tilde{\vec{X}}_{\mathrm{rep}}[i,:] \in \mathbb{R}^{3 \times d_{\mathrm{share}}}$ stacks one token each for the rhythm, morphology, and ST-T repolarization streams of sample $i$. This selective attention ensures $\bar{\vec{z}}_{i,c}^{\mathrm{ecg}}$ and $\bar{\vec{z}}_{i,c}^{\mathrm{rep}}$ isolate patterns specific to label $c$, filtering out co-existing pathological noise and providing a stable foundation for alignment between modalities. ECG keys and values use one token per time step; descriptor keys and values use one token per semantic axis, enabling sub-beat selection on waveforms and axis-specific semantic pooling without collapsing those three axes prematurely.

\subsection{Label Specific Bidirectional Contrastive Learning (LSBC)}
\textbf{LSBC} enforces category-aware consistency within the feature manifolds defined by the LQN outputs. For each diagnostic label $c$, we define the positive sample set $\mathcal{P}_c$ of samples carrying label $c$ and the negative set $\mathcal{N}_c$ of samples without label $c$ in the current batch. The directional InfoNCE-style contrastive loss for label $c$ from modality $\mathrm{m}_1$ to $\mathrm{m}_2$ is:
\vspace{-6pt}
\begin{equation}
\mathcal{L}_{c}^{\mathrm{m}_1 \to \mathrm{m}_2} = - \frac{1}{|\mathcal{P}_c|} \sum_{i \in \mathcal{P}_c} \log \frac{\sum_{j \in \mathcal{P}_c} \exp\left(\mathrm{sim}(\bar{\vec{z}}_{i,c}^{\mathrm{m}_1}, \bar{\vec{z}}_{j,c}^{\mathrm{m}_2})/\tau\right)}{\sum_{k \in \mathcal{P}_c \cup \mathcal{N}_c} \exp\left(\mathrm{sim}(\bar{\vec{z}}_{i,c}^{\mathrm{m}_1}, \bar{\vec{z}}_{k,c}^{\mathrm{m}_2})/\tau\right)}.
\end{equation}
The total LSBC loss is averaged over active labels in the batch $\mathcal{C}_B$ and both alignment directions between modalities:
\vspace{-6pt}
\begin{equation}
\mathcal{L}_{\mathrm{LSBC}} = \frac{1}{|\mathcal{C}_B|}\sum_{c \in \mathcal{C}_B} \frac{1}{2} \left(\mathcal{L}_{c}^{\mathrm{ecg} \to \mathrm{rep}} + \mathcal{L}_{c}^{\mathrm{rep} \to \mathrm{ecg}} \right).
\end{equation}

\subsection{Curriculum Adaptive Fusion (CAF): Gated Optimization}
During transfer, early alignment between modalities can introduce strong contrastive gradients before unimodal encoders have stabilized, disrupting discriminative features acquired during pretraining. CAF acts as a regulator aware of convergence to delay aggressive alignment until training is stable. The dynamic weight of the LSBC loss is jointly governed by normalized training progress $t \in [0,1]$ and a stability gate:
\vspace{-4pt}
\begin{equation}
w_{\mathrm{LSBC},t} = \beta_t \cdot \mathbb{I} \left( \Delta_t < \varepsilon \right),
\end{equation}
where $\mathbb{I}(\cdot)$ is the indicator function and $\varepsilon$ is a predefined stability threshold. The stability gate monitors the exponential moving average (EMA) of the multi-label classification loss to quantify training convergence:
\vspace{-4pt}
\begin{equation}
\begin{split}
\mathrm{EMA}_t &= (1 - \gamma)\mathrm{EMA}_{t-1} + \gamma \mathcal{L}_{\mathrm{ce},t}, \\
\Delta_t &= \left| \mathrm{EMA}_t - \mathrm{EMA}_{t-K} \right|.
\end{split}
\end{equation}
where $\gamma$ is the EMA decay rate, $\mathcal{L}_{\mathrm{ce},t}$ is the instantaneous multi-label classification loss at step $t$, and $K$ is the sliding window size for convergence monitoring.

The alignment intensity is further modulated by a three-stage curriculum weight $\beta_t$ (normalized to $[0,1]$):
\vspace{-6pt}
\begin{equation}
\beta_t =
\begin{cases}
0.1 - 0.1 \cdot \frac{t}{0.3}, & t \in [0, 0.3), \\
0.3 \cdot \frac{t - 0.3}{0.4}, & t \in [0.3, 0.7), \\
0.3 + 0.7 \cdot \frac{t - 0.7}{0.3}, & t \in [0.7, 1].
\end{cases}
\end{equation}
The first curriculum stage therefore keeps the LSBC multiplier small and, under this schedule, decreases it further while the classifier stabilizes; subsequent stages progressively raise $\beta_t$ so that contrastive alignment strengthens only after early optimization has settled.
We additionally tested multiple curriculum breakpoints in the 50-shot regime and observed robust performance: [0.2, 0.6]$\rightarrow$0.782 AUC, [0.3, 0.7]$\rightarrow$0.790 AUC, and [0.4, 0.8]$\rightarrow$0.788 AUC.
\subsection{Final Training Objective}
The entire PEACE framework is jointly optimized by fusing multi-label classification loss and curriculum weighted LSBC contrastive loss. The total training objective is:
\vspace{-4pt}
\begin{equation}
\mathcal{L} = \left(\mathcal{L}_{\mathrm{ce,ecg}} + \mathcal{L}_{\mathrm{ce,rep}}\right) + \lambda_{\max} \cdot w_{\mathrm{LSBC},t} \cdot \mathcal{L}_{\mathrm{LSBC}},
\end{equation}
where $\mathcal{L}_{\mathrm{ce,ecg}}$ and $\mathcal{L}_{\mathrm{ce,rep}}$ denote the multi-label binary cross-entropy losses for the ECG encoder and the semantic descriptor encoder on the original diagnostic label task, and $\lambda_{\max}$ is a hyper-parameter that controls the maximum alignment strength of the LSBC loss. For the semantic branch, descriptors are constructed from training labels and therefore serve as privileged auxiliary supervision during training; this branch is discarded at inference. We report No-text and single-text controls to quantify gains from structured semantic content beyond training on ECG alone.

\section{Experiments}

\subsection{Datasets}
\label{experiment_dataset}
The PEACE model is pretrained on MIMIC-IV \cite{Gow2023MIMICIVECG}, a comprehensive adult ECG dataset comprising over 800,000 records. Evaluation is conducted on two target domains: pediatric ZZU-pECG \cite{2025zzu_pecg}, ages 0 to 14, and PTB-XL \cite{Wagner2020PTBXL}.

\textbf{Preprocessing and harmonization.} All ECGs were standardized to 10-s twelve lead signals at 500 Hz. For MIMIC-IV-ECG, free text diagnostic statements were mapped to the unified twelve label ontology after removing low quality or missing annotation records. For ZZU-pECG, database provided diagnostic codes were converted to the same ontology, and waveforms were filtered using a 0.5 to 100 Hz band-pass filter and a 50 Hz notch filter, followed by amplitude normalization using source-domain training statistics for transfer consistency. For PTB-XL, SCP codes were harmonized to the same ontology and signals were standardized using training set scaling before patient level 8:1:1 splitting.

ZZU-pECG contains 11,643 children in the released database. After filtering and label mapping, we retained 7,593 valid ECG records for experiments, corresponding to 9,198 label instances due to multi-label annotation. Records are partitioned into training, validation, and test folds by patient identifier with an 8:1:1 ratio under multi-label stratification.

For PTB-XL external scoring, SCP codes are harmonized to the same twelve label ontology; under our mapping, nine labels have nonzero mapped incidence (\emph{NORM}, \emph{CRBBB}, \emph{IRBBB}, \emph{LAFB}, \emph{LAO/LAE}, \emph{RAO/RAE}, \emph{LVH}, \emph{RVH}, and \emph{STTC}), totaling 17{,}818 label instances in the processed corpus. \TABabbr, \emph{LVOLT}, and \emph{LQTS} have zero mapped PTB-XL instances and are excluded from the PTB-XL macro average so the headline metric is not averaged over absent heads.

The harmonized cohorts exhibit severe label imbalance and cross-dataset prevalence shift: for example, \TABabbr{} is frequent in ZZU-pECG (2,544 label instances) but has no mapped support in PTB-XL under our harmonization, whereas \emph{STTC} remains represented in all three cohorts. Full names for all diagnostic labels are detailed in the note of Table~\ref{tab:per_class}.

To enable structured semantic supervision, we do not assign each diagnostic label to a single mutually exclusive clinical category. Instead, for every diagnostic label, we construct three axis-specific semantic descriptors corresponding to rhythm, morphology, and ST-T repolarization. This design follows the systematic ECG reading process, where clinicians inspect rate and rhythm organization, waveform morphology and conduction patterns, and repolarization related ST-T and QT findings as complementary evidence sources.

Specifically, the rhythm axis describes rate, regularity, sinus organization, and P-QRS temporal relationships; the morphology axis describes P-wave, QRS, axis, voltage, conduction, and chamber-related findings~\cite{surawicz2009aha}; and the ST-T repolarization axis describes ST-segment, T-wave, and U-wave abnormalities, together with QT or QTc interval abnormalities~\cite{rautaharju2009ecg_iv}. Therefore, each label is represented by a triplet of semantic descriptors rather than by a single text prompt:
\[
\mathcal{D}_c = \{\mathcal{D}_c^{\mathrm{rhythm}}, \mathcal{D}_c^{\mathrm{morph}}, \mathcal{D}_c^{\mathrm{STT}}\},
\]
where $c$ denotes the diagnostic label. These descriptors are used only as privileged semantic supervision during training, while inference remains ECG only.

\subsection{Experimental Settings}
\label{sec:experimental_settings}

\textbf{Architecture and Optimization.} PEACE is instantiated using a ResNet1D backbone \cite{resnetwang} for waveform encoding and BioClinicalBERT for clinical text processing. To balance knowledge retention with domain adaptation, we employ a graduated freezing strategy where the first 10 transformer blocks of BioClinicalBERT remain fixed. The model is optimized using AdamW with a cosine annealing learning rate scheduler. To address the long-tailed distribution of clinical labels, we incorporate class-balanced loss weighting.

\textbf{Optimization hyperparameters.} Unless stated otherwise for the transfer schedules below, we use AdamW with $\beta_1 = 0.9$, $\beta_2 = 0.999$, weight decay $1.2 \times 10^{-3}$, initial learning rate $4 \times 10^{-4}$, 5-epoch linear warmup from $1 \times 10^{-5}$, cosine decay to $1 \times 10^{-6}$ over the remaining 95 epochs, batch size $1024$, gradient clipping at 5.0, and inverse frequency class weights for multi-label BCE clipped to $[1.0,5.0]$. MIMIC-IV pretraining runs for 100 epochs; the first ten BioClinicalBERT transformer blocks remain frozen while higher layers, the pooler, and projection heads train, together with the ResNet1D encoder and LQN fusion.

\textbf{Transfer Learning Regimes.} We evaluate PEACE under three clinical deployment scenarios.
\textbf{Zero-shot Transfer:} The pretrained model remains frozen; AUC is computed from prediction scores, while validation-selected thresholds are used only for threshold-dependent metrics.
\textbf{Few-shot Adaptation:} The model is fine-tuned on 50 labeled instances per class (or the maximum available for rare classes) for 20 epochs at learning rate $2.5 \times 10^{-5}$ with the same freezing policy; this $N=50$ setting is supported by the sample-efficiency analysis in Table~\ref{tab:sample_efficiency} and Figure~\ref{fig:sample_efficiency}.
\textbf{Full Supervised Fine-tuning:} Training uses the full labeled training split for 20 epochs at $1 \times 10^{-4}$ with the same text encoder freezing policy, and we keep the checkpoint with highest macro-average validation AUC.

\paragraph{Baseline models}

To evaluate PEACE from both ECG representation learning and multimodal fusion perspectives, we organize the baselines into two groups.

\textbf{Group A: Domain adaptation and standard fusion baselines.} This group evaluates whether conventional distribution alignment or simple multimodal fusion is sufficient under adult-to-pediatric transfer. DANN \cite{ganin2016domain} represents adversarial domain adaptation, and MMD \cite{long2015learning} represents distribution matching in feature space. Both methods use the same ResNet1D backbone initialization as PEACE. In the zero-shot setting, DANN and MMD are evaluated without accessing any ZZU-pECG target-domain samples, either labeled or unlabeled. Their near-random zero-shot AUCs are therefore consistent with the fact that conventional domain adaptation methods typically require target-domain data to estimate or align the target distribution. For the main few-shot and full fine-tuning comparisons, we use a labels-only BCE protocol on ZZU-pECG with the same fine-tuning budget as PEACE. This choice isolates the effect of model design under an identical pediatric supervision regime rather than introducing additional unlabeled-source/target mixing that would change the transfer budget. 

We also include two standard ECG--text fusion baselines. Early-fusion concatenates the ECG embedding produced by ResNet1D with the BioClinicalBERT descriptor embedding and feeds the joint vector into an MLP classifier. Late-fusion trains ECG and text branches separately and averages their logits at the decision level. These two baselines use the same ECG encoder, clinical BERT encoder, and class-level descriptor source as PEACE, but remove PEACE's label-query interaction, label-specific contrastive alignment, and curriculum-adaptive fusion.

\textbf{Group B: ECG foundation and semantic pretraining baselines.} ST-MEM \cite{STMEM} provides a masked ECG representation learning reference. MERL \cite{MERL} evaluates global multimodal alignment with test-time clinical knowledge enhancement. KED \cite{2024KED} evaluates knowledge-enhanced ECG representation learning with textual-signal alignment. Together, these baselines cover masked waveform pretraining, global ECG--text alignment, and knowledge-enhanced representation learning.

This grouping is intended to answer two complementary questions: whether PEACE improves over conventional domain adaptation and naive fusion strategies, and whether its label-conditioned fusion mechanism improves over existing semantic or foundation-model pretraining approaches.

\textbf{Evaluation metrics.} We adopt macro-average area under the ROC curve (macro-average AUC) as the primary metric; it summarizes ranking performance across operating points and is computed directly from prediction scores without committing to a decision threshold. We additionally report macro balanced accuracy (BAcc), defined as the macro average of $(\mathrm{Sensitivity}+\mathrm{Specificity})/2$ per label, and macro-F1 using validation-tuned per-class thresholds (Section~\ref{sec:experimental_settings}). Table~\ref{tab:main_results} summarizes these metrics together with external PTB-XL macro-average AUC after full fine-tuning.

\paragraph{Evaluation protocol and reproducibility} Unless stated otherwise, stochastic experiments were repeated with five random seeds, $\{42,43,44,45,46\}$. We use patient-level train/validation/test splits at an 8:1:1 ratio on ZZU-pECG and PTB-XL. Checkpoints are selected according to the highest macro-average validation AUC. For threshold-dependent metrics, including macro-F1, specificity, and balanced accuracy, per-class decision thresholds are tuned on the validation split to maximize validation macro-F1 and are then fixed for test evaluation. Macro-average AUC is computed directly from prediction scores and is used as the primary threshold-free metric. Optimization schedules for pretraining and transfer follow Section 4.2. The complete raw-code-to-label harmonization map and generated semantic descriptors will be released with the code upon acceptance to facilitate reproducibility.

\begin{table}[t] 
\centering 
\caption{Main performance comparison on ZZU-pECG and external PTB-XL validation. Group A contains domain-adaptation and standard fusion baselines without label-conditioned semantic alignment. Group B contains ECG foundation or semantic pretraining baselines. PEACE is shown separately as the proposed label-conditioned multimodal fusion framework.} 
\label{tab:main_results} 
\resizebox{\textwidth}{!}{ \begin{tabular}{lcccccccccc} \toprule \multirow{2}{*}{Model} & \multicolumn{3}{c}{Zero-shot} & \multicolumn{3}{c}{50-shot} & \multicolumn{3}{c}{Full FT} & \multirow{2}{*}{\makecell{PTB-XL Full FT\\AUC}} \\ \cmidrule(lr){2-4} \cmidrule(lr){5-7} \cmidrule(lr){8-10} & AUC & BAcc & F1 & AUC & BAcc & F1 & AUC & BAcc & F1 & \\ \midrule \multicolumn{11}{l}{\textit{Group A --- Domain adaptation and standard fusion baselines without label-conditioned semantics}} \\ \midrule DANN \cite{ganin2016domain} & 49.33 & 54.89 & 20.56 & \underline{81.70} & \underline{72.14} & \underline{48.97} & 89.67 & 79.05 & 58.72 & 96.54 \\ MMD \cite{long2015learning} & 49.33 & 54.89 & 20.56 & 80.62 & 71.41 & 46.67 & \underline{89.94} & \underline{79.92} & \underline{59.57} & 96.51 \\ Early-fusion & \underline{51.88} & \underline{55.85} & \underline{20.87} & 80.69 & 72.02 & 44.77 & 88.87 & 78.03 & 57.07 & \underline{96.69} \\ Late-fusion & 50.55 & 55.32 & 19.58 & 78.22 & 69.84 & 42.41 & 89.03 & 79.48 & 56.77 & 96.60 \\ \midrule \multicolumn{11}{l}{\textit{Group B --- ECG foundation and semantic pretraining baselines}} \\ \midrule ST-MEM \cite{STMEM} & 50.91 & 51.70 & 15.54 & 66.08 & 58.86 & 23.86 & \underline{81.55} & \underline{66.60} & \underline{88.85} & 89.22 \\ MERL \cite{MERL} & \underline{58.58} & 51.34 & 4.57 & \underline{70.65} & \underline{59.65} & \underline{26.35} & 80.97 & 65.94 & 33.72 & \underline{92.49} \\ KED \cite{2024KED} & 57.77 & 
\underline{53.57} & \underline{17.30} & 50.66 & 51.10 & 14.97 & 80.54 & 63.91 & 25.97 & 89.10 \\ \midrule \multicolumn{11}{l}{\textit{Proposed label-conditioned multimodal fusion framework}} \\ 
\midrule \rowcolor{gray!10} \textbf{PEACE} & \textbf{59.39} & 52.30 & 17.30 & \textbf{81.74} & 69.52 & 37.59 & \textbf{91.56} & \textbf{80.50} & 59.08 & \textbf{96.90} \\ 
\rowcolor{gray!5} $\Delta_{\mathrm{AUC}}$ vs. Group A best & \textbf{+7.51} & -- & -- & \textbf{+0.04} & -- & -- & \textbf{+1.62} & -- & -- & \textbf{+0.21} \\ 
\rowcolor{gray!5} $\Delta_{\mathrm{AUC}}$ vs. Group B best & \textbf{+0.81} & -- & -- & \textbf{+11.09} & -- & -- & \textbf{+10.01} & -- & -- & \textbf{+4.41} \\ 
\bottomrule 
\end{tabular} } 
\vspace{-0.5em} 
\begin{flushleft} \footnotesize BAcc = (Sensitivity + Specificity) / 2. All values are macro averages in \%. Underlined values indicate the best result within each baseline group. Bold values indicate the best overall result or a positive PEACE gain in the primary AUC metric. $\Delta_{\mathrm{AUC}}$ denotes the absolute AUC difference between PEACE and the strongest baseline in the corresponding group and evaluation regime. All stochastic experiments were repeated with five random seeds: 42, 43, 44, 45, and 46. 
\end{flushleft} 
\end{table}

\subsection{Results and Analyses} 

We consolidate cross-domain pediatric transfer metrics on ZZU-pECG under zero-shot, 50-shot, and full fine-tuning settings, together with external PTB-XL validation after full fine-tuning, in Table~\ref{tab:main_results}. The baselines are grouped to distinguish conventional domain adaptation and standard fusion baselines from ECG foundation or semantic pretraining baselines. 

Table~\ref{tab:main_results} is designed as a controlled comparison across three alternatives: generic domain adaptation, standard ECG--text fusion, and existing ECG foundation or semantic pretraining baselines. PEACE achieves the best macro-AUC across the primary ZZU-pECG transfer regimes, with the largest gains appearing under 50-shot and full fine-tuning against Group B. This pattern indicates that the proposed fusion strategy is most beneficial when limited pediatric supervision is available, rather than merely improving adult-domain supervised validation.

\paragraph{Comparison with domain adaptation and standard fusion baselines} Group A evaluates whether distribution alignment or simple ECG--text fusion is sufficient without label-conditioned semantic alignment. All four baselines in this group obtain zero-shot AUCs between 49.33\% and 51.88\% on ZZU-pECG, whereas PEACE reaches 59.39\%, yielding a 7.51 percentage-point gain over the strongest Group A baseline. For DANN and MMD, this zero-shot setting does not expose the model to unlabeled ZZU-pECG target-domain samples; therefore, the domain-adaptation objective cannot estimate the target distribution. Their near-random zero-shot AUCs are consistent with this protocol and with the usual requirement of domain-adaptation methods for target-domain data during alignment. 

Under 50-shot adaptation, PEACE reaches 81.74\% AUC, slightly exceeding DANN, the strongest Group A method, by 0.04 percentage points. DANN obtains higher threshold-dependent BAcc and F1 in this regime, which indicates that validation-tuned thresholds can favor a particular operating point under severe label imbalance. In contrast, PEACE remains strongest in the primary threshold-free AUC metric and achieves the best full fine-tuning BAcc, suggesting a more balanced ranking and sensitivity--specificity trade-off after sufficient pediatric adaptation. Under full fine-tuning, PEACE achieves 91.56\% AUC and 80.50\% BAcc, outperforming the best Group A results by 1.62 and 0.58 percentage points, respectively. 

PTB-XL AUC after full fine-tuning is close across Group A baselines and PEACE (96.51--96.90\%). This convergence is expected because PTB-XL provides relatively abundant adult-domain supervision, under which conventional supervised fine-tuning can already recover strong in-domain performance. Thus, the main evidence from Group A lies in pediatric transfer, especially the zero-shot initialization gap and the full fine-tuning improvement on ZZU-pECG.

\paragraph{Comparison with ECG foundation and semantic pretraining baselines} 

PEACE's advantage is more pronounced against Group B. At zero-shot, PEACE achieves 59.39\% AUC, modestly surpassing MERL (58.58\%) and KED (57.77\%), and substantially exceeding ST-MEM (50.91\%). The gap widens sharply under 50-shot adaptation: PEACE reaches 81.74\% AUC, compared with 70.65\% for MERL, 66.08\% for ST-MEM, and 50.66\% for KED. The 11.09 percentage-point gain over the strongest Group B baseline indicates that label-conditioned alignment and curriculum-adaptive fusion are especially beneficial when pediatric labels are scarce. Under full fine-tuning, PEACE obtains 91.56\% AUC, outperforming the strongest Group B baseline by 10.01 percentage points. On PTB-XL, PEACE reaches 96.90\% AUC, compared with 92.49\% for MERL, 89.22\% for ST-MEM, and 89.10\% for KED. These results suggest that PEACE does not merely improve pediatric adaptation, but also preserves transferable ECG representations under adult-domain validation. Overall, Group B demonstrates the methodological value of structured label-conditioned fusion over global multimodal alignment, masked ECG pretraining, or knowledge distillation alone. 

\paragraph{Implication for multimodal fusion} 

The contrast between PEACE and the Early-/Late-fusion baselines is particularly important from a multimodal fusion perspective. All three methods use ECG and text-derived semantic information, but they differ in how the modalities interact. Early-fusion performs feature concatenation, Late-fusion combines independent modality logits, whereas PEACE uses diagnostic labels as queries to extract label-specific ECG and semantic features before bidirectional contrastive alignment. The consistent AUC gains of PEACE over these fusion baselines indicate that the benefit is not simply due to adding text, but to the proposed label-conditioned interaction and curriculum-gated alignment mechanism.

\begin{table}[t] 
\centering 
\caption{Few-shot macro-average AUC (\%) on ZZU-pECG vs. shots per class $N$. Row-wise $\Delta$ and relative improvement are computed against the preceding row.} 
\label{tab:sample_efficiency} 
\begin{tabular}{lccc} 
\toprule Configuration & AUC (mean$\pm$std) & $\Delta$ gain & Rel. imp. \\ \midrule 5-shot & 64.85$\pm$4.48 & -- & -- \\ 10-shot & 66.13$\pm$3.33 & +1.28 & +1.97\% \\ 20-shot & 73.10$\pm$1.64 & +6.97 & +10.54\% \\ 50-shot & 81.74$\pm$1.34 & \textbf{+8.64} & \textbf{+11.82\%} \\ 100-shot & 84.53$\pm$1.17 & +2.79 & +3.41\% \\ 
\bottomrule 
\end{tabular} 
\end{table}

\begin{figure}[t] 
\centering 
\includegraphics[width=0.7\linewidth]{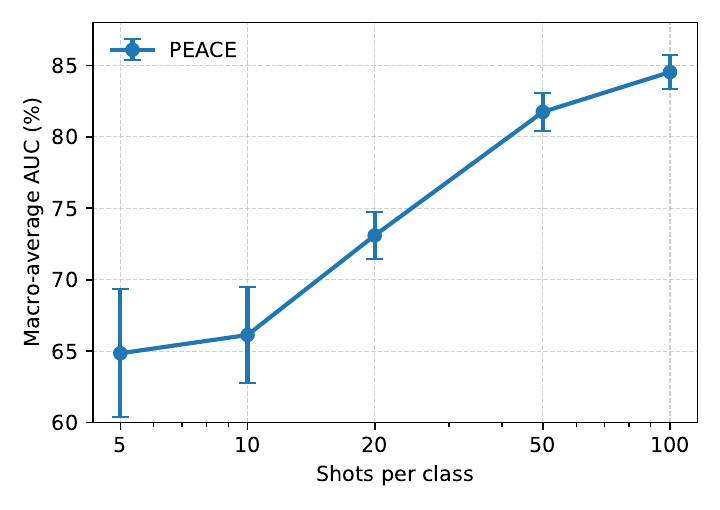} 
\caption{Few-shot sample efficiency on ZZU-pECG. PEACE improves rapidly from 10-shot to 50-shot and begins to saturate beyond 50 labeled examples per class, suggesting that the proposed label-conditioned fusion mechanism is most beneficial under low-resource pediatric adaptation.} 
\label{fig:sample_efficiency} 
\end{figure}

\paragraph{Few-shot sample efficiency} Figure~\ref{fig:sample_efficiency} visualizes the sample-efficiency trend reported in Table~\ref{tab:sample_efficiency}. PEACE exhibits three regimes: a cold-start regime below 10 shots, rapid adaptation between 10 and 50 shots, and saturation beyond 50 shots. The gain from 20-shot to 50-shot is substantially larger than that from 50-shot to 100-shot, indicating that 50-shot provides a practical annotation budget that captures most of the benefit of pediatric supervision.

\begin{table}[t] 
\centering 
\caption{Ablation on ZZU-pECG: macro-average AUC (\%, mean$\pm$std over five seeds). 
Single-text uses one fused descriptor per label; Full PEACE uses separate rhythm, morphology, and ST-T repolarization descriptors.}
\label{tab:ablation} 
\begin{tabular}{lccc} 
\toprule 
Configuration & Zero-shot & 50-shot & Full FT \\ 
\midrule 
No-text 
& 46.01$\pm$0.00 
& 66.37$\pm$1.80 
& 66.86$\pm$3.89 \\ 

w/o LSBC 
& 56.23$\pm$0.00 
& 77.74$\pm$0.33 
& 89.63$\pm$0.86 \\

w/o CAF 
& 47.10$\pm$0.00 
& 77.69$\pm$0.81 
& 90.45$\pm$0.35 \\ 

Single-text PEACE 
& 49.20$\pm$3.00 
& 74.30$\pm$1.50 
& 88.70$\pm$1.10 \\ 

\rowcolor{gray!10} 
Full PEACE 
& \textbf{59.39$\pm$0.00} 
& \textbf{81.74$\pm$1.34} 
& \textbf{91.56$\pm$0.96} \\ 
\bottomrule 
\end{tabular} 
\end{table}

\subsection{Ablation Studies}

Table~\ref{tab:ablation} reports module ablations and semantic supervision controls on ZZU-pECG. 
The ablations are designed to answer three fusion-specific questions: 
(i) whether semantic supervision contributes beyond ECG-only training; 
(ii) whether label-specific bidirectional alignment is necessary beyond standard classification; and 
(iii) whether the structured three-axis descriptor design is more effective than a single fused text prompt.

The No-text configuration removes the semantic descriptor branch while retaining the remaining PEACE training recipe, serving as an internal control for the role of semantic supervision. 
Removing LSBC degrades performance across all regimes, confirming that label-specific bidirectional ECG--semantic alignment contributes beyond unimodal classification. 
Removing CAF is particularly harmful under zero-shot and few-shot transfer, indicating that aggressive cross-modal alignment can destabilize pretrained ECG representations when target labels are scarce. 
The Single-text PEACE variant uses one fused descriptor per class, whereas full PEACE separates rhythm, morphology, and ST-T repolarization descriptors. 
Full PEACE consistently outperforms Single-text PEACE across zero-shot, 50-shot, and full fine-tuning settings, suggesting that structured decomposition of diagnostic semantics can reduce prompt-level entanglement and improve transferability compared with a single undifferentiated descriptor.

We do not assign each label to a single semantic axis or report axis-only diagnostic performance, because rhythm, morphology, and ST-T repolarization are not independent modalities or mutually exclusive disease groups. 
Therefore, we use the single-text variant as the controlled baseline to test whether structured decomposition is preferable to an undifferentiated label descriptor.

Table~\ref{tab:ablation} reports the contribution of semantic supervision, LSBC, CAF, and descriptor structure. Full PEACE achieves the highest AUC across zero-shot, 50-shot, and full fine-tuning settings. Removing the semantic branch produces the largest degradation, indicating that ECG-only training under the same optimization protocol cannot recover the full transfer benefit. Removing LSBC or CAF consistently lowers AUC, with the effect most visible under zero-shot and few-shot transfer. Replacing the three-axis descriptors with a single fused descriptor also reduces performance across all regimes. These results show that the observed gains depend on the combined use of structured semantic descriptors, label-level alignment, and curriculum-gated fusion rather than on a single component alone.

\subsection{Interpretability Analyses}
\label{sec:interpretability}

\begin{figure}[!t]
\centering
\includegraphics[width=\linewidth]{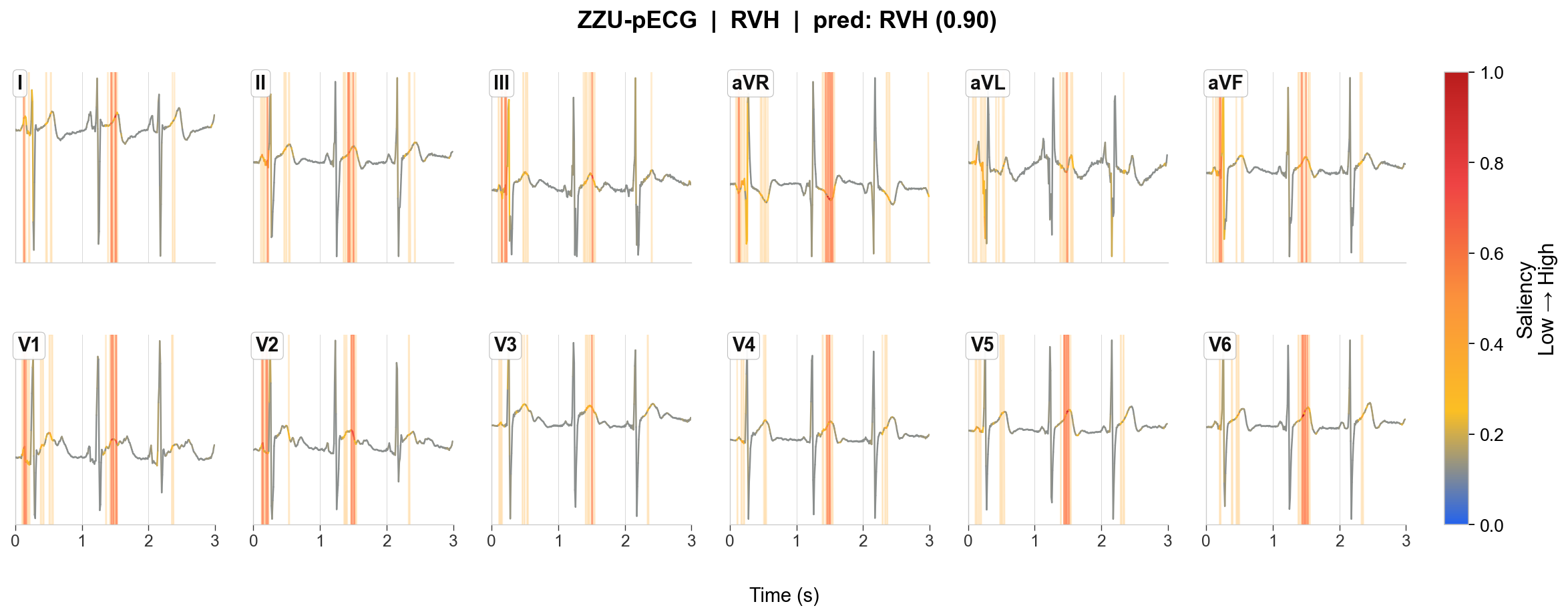}
\caption{Grad-CAM++ saliency for RVH on a representative pediatric ZZU-pECG recording. Gray curves denote ECG waveforms, and warm-colored overlays indicate normalized saliency intensity. Saliency is mainly concentrated around QRS-dominant voltage and morphology regions across right-precordial and limb leads, consistent with ECG evidence commonly inspected for ventricular hypertrophy.}
\label{FIG:zzu_RVH_leads}
\end{figure}

\begin{figure}[!t]
\centering
\includegraphics[width=\linewidth]{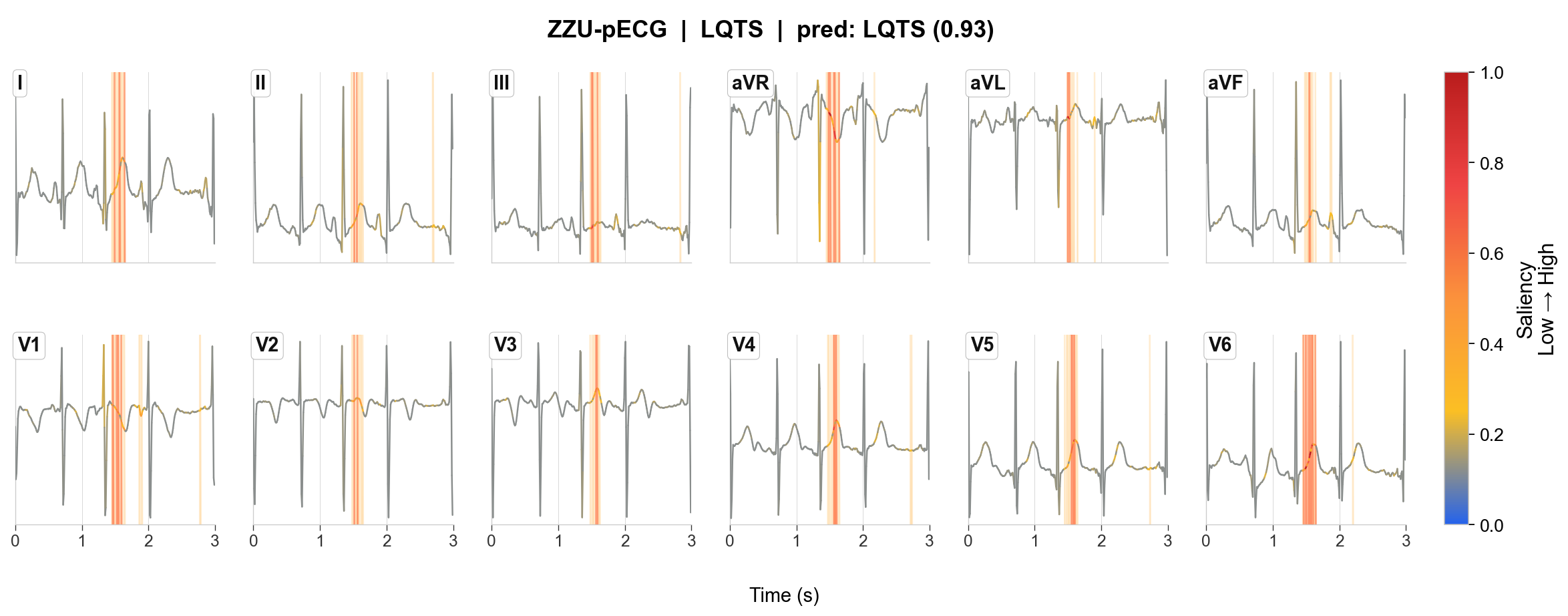}
\caption{Grad-CAM++ saliency for LQTS on a representative pediatric ZZU-pECG recording. Gray curves denote ECG waveforms, and warm-colored overlays indicate normalized saliency intensity. Saliency is concentrated around QRS-to-T and repolarization-related intervals across multiple leads, broadly aligning with QT/ST-T evidence used in ECG interpretation.}
\label{FIG:zzu_LQTS_leads}
\end{figure}

To verify whether model decisions emphasize clinically plausible waveform regions, we visualize Grad-CAM++ \cite{gradcamplus} attributions on representative pediatric ZZU-pECG recordings using the same preprocessing and checkpoint selection as Section~4.2. Figure~\ref{FIG:zzu_RVH_leads} shows an RVH example, where saliency is mainly concentrated around QRS-dominant voltage and morphology regions across right-precordial and limb leads. This pattern is broadly consistent with ECG evidence commonly inspected for ventricular hypertrophy and chamber-related abnormalities. Figure~\ref{FIG:zzu_LQTS_leads} shows an LQTS example, where saliency shows increased activation around QRS-to-T and repolarization-related intervals across multiple leads, aligning with QT/ST-T evidence used in ECG interpretation.

Across both examples, attributions remain temporally localized rather than uniformly distributed across the full tracing, suggesting that the label-query mechanism encourages attention to label-relevant intervals instead of relying solely on global pooled features. These maps are qualitative single-record visualizations and do not constitute causal explanations or standalone diagnostic evidence.

\subsection{Threshold-dependent metrics under severe imbalance}
\label{sec:auc_f1_discussion}
Because macro-AUC is threshold-free whereas BAcc and macro-F1 depend on validation-selected operating points, these metrics may rank models differently under severe multi-label imbalance. In Table~\ref{tab:main_results}, DANN obtains higher 50-shot BAcc and F1 despite a slightly lower AUC than PEACE, while ST-MEM obtains a higher full fine-tuning macro-F1 but substantially lower macro-AUC and BAcc. These discrepancies suggest that threshold-dependent metrics can reflect favorable operating-point calibration rather than uniformly better ranking performance. We therefore use macro-AUC as the primary metric and report BAcc and macro-F1 as complementary measures of threshold-specific behavior.

\section{Discussion}

\subsection{PTB-XL convergence under abundant adult supervision}

The PTB-XL AUC convergence among Group A baselines and PEACE after full fine-tuning is consistent with the relatively abundant adult supervision in that regime. When sufficient target-domain adult labels are available, conventional supervised fine-tuning can already recover strong in-domain performance. The structural advantage of label-conditioned semantic fusion is therefore most evident under pediatric domain shift and label scarcity, particularly in zero-shot and 50-shot ZZU-pECG transfer, rather than in fully supervised adult-domain validation.

\subsection{Cross-population findings and practical scope}

Table~\ref{tab:per_class} complements Table~\ref{tab:main_results} with per-label trajectories across evaluation regimes. PEACE improves consistently as pediatric supervision grows, and Table~\ref{tab:ablation} indicates that CAF is most consequential when labels are scarce. Macro-F1 and macro-AUC need not rank models identically under imbalance, as discussed in Section~4.6. Age-stratified evaluation by neonatal, infant, preschool, and school-age groups remains an important next step because pediatric ECG criteria vary substantially across developmental stages. 

Without pediatric updates, PEACE reaches 59.39\% macro-average AUC at zero-shot, already exceeding several evaluated baselines, and then climbs to 81.74\% with only 50 labeled examples per class. This trajectory supports the role of CAF in stabilizing low-resource pediatric adaptation. The weakest zero-shot categories, including NORM and LVH, are also those most sensitive to pediatric heart-rate and voltage norms; PEACE should therefore be viewed as a transferable initialization that becomes clinically meaningful after limited pediatric supervision, rather than as a standalone zero-shot pediatric classifier. 

Differentiating benign developmental variants from pathology is a central pediatric challenge, especially for chamber-related morphology, voltage criteria, conduction patterns, and repolarization intervals. The RVH and LQTS saliency examples in Section~4.5 are consistent with this interpretation, as the model shows increased activation around QRS-dominant morphology for RVH and QRS-to-T/repolarization intervals for LQTS. IRBBB often reflects normal maturation; PEACE starts at 83.82\% zero-shot AUC on this class and rises to 96.60\% after full fine-tuning. Ventricular hypertrophy-related labels similarly stress voltage and chamber-related morphology criteria shared with adult models; PEACE improves LVH from 35.93\% to 90.23\% AUC and RVH from 55.51\% to 97.67\% AUC after full fine-tuning. For rare or clinically subtle entities such as LQTS, PEACE improves from 51.61\% to 87.21\% AUC, suggesting that label-conditioned alignment helps reduce collapse of tail classes into dominant clusters. These heterogeneous mechanisms, including developmental right-heart dominance, age-dependent rate decline, voltage remodeling, and repolarization-interval variation, interact nonlinearly; PEACE addresses this multiplicity through structured semantics and gated fusion rather than a single scalar age adjustment.

\begin{table}[ht]
\centering
\setlength{\tabcolsep}{4pt} 
\footnotesize
\caption{Per-class performance comparison of four frameworks on ZZU-pECG across three evaluation regimes. ``FT'' denotes fine-tuning. Best performance in each regime is highlighted in \textbf{bold} with a \colorbox{gray!20}{gray background}.}
\resizebox{\textwidth}{!}{
\begin{tabular}{
>{\raggedright\arraybackslash}p{1.6cm} 
*{3}{>{\centering\arraybackslash}p{1.2cm} >{\centering\arraybackslash}p{0.9cm} >{\centering\arraybackslash}p{0.9cm} >{\centering\arraybackslash}p{1.5cm}}
}
\toprule 
\multirow{2}{*}{\textbf{Label}} & \multicolumn{4}{c}{\textbf{Zero-shot}} & \multicolumn{4}{c}{\textbf{50-shot}} & \multicolumn{4}{c}{\textbf{Full FT}} \\
\cmidrule(lr){2-5} \cmidrule(lr){6-9} \cmidrule(lr){10-13}
 & PEACE & KED & MERL & ST-MEM & PEACE & KED & MERL & ST-MEM & PEACE & KED & MERL & ST-MEM \\
\midrule 
CRBBB & 67.86 & 47.14 & \cellcolor{gray!20}\textbf{87.42} & 48.86 & \cellcolor{gray!20}\textbf{96.97} & 55.89 & 94.68 & 89.59 & \cellcolor{gray!20}\textbf{99.94} & 94.46 & 98.21 & 97.08 \\
IRBBB & \cellcolor{gray!20}\textbf{83.82} & 56.89 & 58.25 & 49.18 & \cellcolor{gray!20}\textbf{90.78} & 60.18 & 65.61 & 60.67 & \cellcolor{gray!20}\textbf{96.60} & 87.12 & 81.00 & 82.38 \\
LAFB & \cellcolor{gray!20}\textbf{69.60} & 46.40 & 66.01 & 56.68 & \cellcolor{gray!20}\textbf{95.28} & 56.76 & 76.11 & 82.04 & \cellcolor{gray!20}\textbf{99.31} & 87.14 & 89.44 & 94.36 \\
LAO/LAE & \cellcolor{gray!20}\textbf{66.65} & 61.02 & 44.31 & 50.79 & \cellcolor{gray!20}\textbf{79.23} & 51.28 & 63.58 & 50.00 & \cellcolor{gray!20}\textbf{92.85} & 78.54 & 69.24 & 67.97 \\
LQTS & 51.61 & 53.48 & \cellcolor{gray!20}\textbf{64.71} & 51.34 & \cellcolor{gray!20}\textbf{76.03} & 55.27 & 60.87 & 67.61 & \cellcolor{gray!20}\textbf{87.21} & 70.37 & 77.82 & 81.80 \\
LVH & 35.93 & 48.08 & 37.97 & \cellcolor{gray!20}\textbf{50.32} & \cellcolor{gray!20}\textbf{80.93} & 37.95 & 71.70 & 59.29 & \cellcolor{gray!20}\textbf{90.23} & 75.07 & 74.48 & 71.89 \\
LVOLT & 60.53 & \cellcolor{gray!20}\textbf{66.65} & 64.04 & 53.60 & \cellcolor{gray!20}\textbf{71.04} & 41.97 & 56.48 & 56.64 & \cellcolor{gray!20}\textbf{86.33} & 74.73 & 73.61 & 75.93 \\
NORM & 46.16 & \cellcolor{gray!20}\textbf{52.90} & 48.83 & 47.32 & \cellcolor{gray!20}\textbf{80.04} & 53.51 & 73.08 & 70.79 & \cellcolor{gray!20}\textbf{89.90} & 80.90 & 82.09 & 80.42 \\
RAO/RAE & 59.67 & \cellcolor{gray!20}\textbf{75.46} & 56.80 & 46.64 & \cellcolor{gray!20}\textbf{86.45} & 43.35 & 76.75 & 56.59 & \cellcolor{gray!20}\textbf{87.10} & 84.13 & 84.89 & 82.00 \\
RVH & 55.51 & 73.03 & \cellcolor{gray!20}\textbf{73.55} & 49.32 & \cellcolor{gray!20}\textbf{85.72} & 47.49 & 82.68 & 69.88 & \cellcolor{gray!20}\textbf{97.67} & 91.44 & 91.73 & 90.79 \\
STTC & 58.45 & 52.43 & 51.77 & \cellcolor{gray!20}\textbf{55.67} & \cellcolor{gray!20}\textbf{77.80} & 49.62 & 68.13 & 74.08 & \cellcolor{gray!20}\textbf{86.80} & 74.21 & 78.85 & 83.68 \\
\TABabbr & 56.90 & \cellcolor{gray!20}\textbf{59.91} & 49.28 & 51.22 & \cellcolor{gray!20}\textbf{60.63} & 54.66 & 58.18 & 55.77 & \cellcolor{gray!20}\textbf{84.83} & 68.35 & 70.23 & 70.33 \\
\midrule
\textbf{Avg.} & \cellcolor{gray!20}\textbf{59.39} & 57.77 & 58.58 & 50.91 & \cellcolor{gray!20}\textbf{81.74} & 50.66 & 70.65 & 66.08 & \cellcolor{gray!20}\textbf{91.56} & 80.54 & 80.97 & 81.55 \\
\bottomrule 
\addlinespace[1ex] 
\multicolumn{13}{p{18.5cm}}{\small \textit{Note:} All results are presented as AUC (\%). Best results in each evaluation regime are highlighted in gray. CRBBB: complete right bundle branch block; IRBBB: incomplete right bundle branch block; LAFB: left anterior fascicular block; LAO/LAE: left atrial enlargement; LQTS: long QT syndrome; LVH: left ventricular hypertrophy; LVOLT: low QRS voltage; NORM: normal ECG; RAO/RAE: right atrial enlargement; RVH: right ventricular hypertrophy; STTC: ST/T changes; TAB\_: T-wave abnormality.}
\end{tabular}
} 
\label{tab:per_class}
\end{table}


Per-label results in Table~\ref{tab:per_class} further show that PEACE improves over the strongest non-PEACE baseline across most pediatric diagnostic categories, with particularly clear gains on morphology- and conduction-sensitive labels such as IRBBB, LAFB, LVH, and RVH. These class-level trends complement the macro-average results in Table~\ref{tab:main_results} and suggest that the benefit of label-conditioned alignment is not restricted to a single high-prevalence label.

\subsection{Limitations}
First, Gemini descriptors are label conditioned synthetic text; although ablations show label identity alone cannot explain the gains, residual stylistic correlations may remain. Some textual norms (e.g., sinus rate wording) skew toward adult-oriented norms and may be suboptimal across pediatric age bands. Second, PEACE performs static label alignment without explicit developmental stage modeling, which may limit robustness during rapid neonatal physiological change. Third, all results are retrospective on curated datasets; model outputs are multi-label classification probabilities (or scores from which AUC is computed), not clinical recommendations, treatment decisions, or guideline concordant diagnoses. Generalization to routine workflows, acquisition hardware, institutions, and preprocessing pipelines beyond those represented here has not been established. Prospective deployment would require expert in the loop validation, calibration, and subgroup monitoring before clinical use. Fourth, continuous age-aware transfer remains future work.

\section{Conclusion}

This work presented PEACE, a label-conditioned cross-modal fusion framework for adult-to-pediatric ECG transfer under limited pediatric supervision. PEACE uses adult-scale ECG pretraining as the physiological initialization and introduces rhythm, morphology, and ST-T repolarization descriptors as privileged semantic supervision during training. Unlike conventional ECG--text fusion, PEACE does not require text input at deployment. Instead, diagnostic labels serve as fusion queries, label-specific bidirectional contrastive learning aligns ECG and semantic representations at the label level, and curriculum-adaptive gating regulates cross-modal alignment during optimization.

Experiments on ZZU-pECG demonstrate that this training-time fusion strategy provides a stronger transferable initialization and better pediatric adaptation than conventional alternatives. PEACE achieves macro-average AUCs of 59.39\%, 81.74\%, and 91.56\% under zero-shot, 50-shot, and full fine-tuning settings, respectively. It outperforms generic domain-adaptation and standard fusion baselines, including DANN, MMD, Early-fusion, and Late-fusion, indicating that distribution matching or simple ECG--text fusion is insufficient for pediatric transfer. Compared with ECG foundation and semantic pretraining baselines, including ST-MEM, MERL, and KED, PEACE shows larger gains under 50-shot and full fine-tuning regimes. Ablation results further support the value of structured semantic descriptors, label-specific alignment, and curriculum-gated fusion.

PEACE also reaches 96.90\% macro-average AUC on PTB-XL after full fine-tuning over nine harmonized labels with nonzero mapped incidence, suggesting that the learned representations remain competitive under adult-domain validation. Nevertheless, PEACE remains a retrospective classification framework rather than a deployable diagnostic system. Future work will focus on developmental stage-aware modeling, age-stratified calibration, prospective multi-center validation, and integration with real pediatric ECG reports when paired waveform-text data become available.

\section*{Declaration of Generative AI and AI-assisted Technologies in the Manuscript Preparation Process}

During the preparation of this work, the authors used generative AI tools to assist with language polishing and organization of the manuscript. Gemini was also used as part of the research design to generate label-level semantic descriptors, as described in Section 3.2 and Appendix A. After using these tools, the authors reviewed, edited, and verified the content as needed and take full responsibility for the content of the published article.

\section*{Data Availability}

MIMIC-IV-ECG and PTB-XL are publicly available subject to their respective data access requirements. ZZU-pECG is available from its original data repository or data provider according to the dataset access policy. The processed label harmonization maps and generated label-level semantic descriptors will be released with the code upon acceptance.

\section*{Acknowledgments}
This work was supported in part by the National Natural Science Foundation of China under Grant 62571123; in part by the Basic Research Program of Jiangsu Province under Grant BK20252010; in part by the Zhi-Shan Young Scholar Program of Southeast University.

\bibliographystyle{elsarticle-num}
\bibliography{custom}

\clearpage

\appendix
\section{Label-Specific Semantic Descriptors}
\label{app:report_gemini}

The following label-level semantic descriptors were used as privileged auxiliary supervision during training. They are not patient-specific diagnostic reports and are not intended to serve as pediatric diagnostic criteria.

\begin{itemize}
\item \textbf{CRBBB}: rhythm: Regular sinus rhythm showing stable atrioventricular conduction; morphology: QRS $\geq$ 120 ms with characteristic rsR' or rSR' in right precordial leads V1-V2 and slurred terminal S wave in I and V6; ST-T repolarization: Secondary T-wave inversion in V1-V3 without territorial ST-segment elevation or reciprocal changes.
\item \textbf{IRBBB}: rhythm: Normal sinus rhythm with consistent R-R intervals; morphology: QRS duration 100-119 ms exhibiting rSr' or rsR' in V1-V2 and narrow terminal S in I and V6; ST-T repolarization: Absence of acute ST-T changes; no pathologic Q waves or ST-segment deviation in contiguous leads.
\item \textbf{LAFB}: rhythm: Sinus rhythm with normal heart rate; morphology: Marked left axis deviation $-45$ to $-90$ degrees, qR pattern in lateral leads I and aVL, and rS pattern in inferior leads II, III, aVF; ST-T repolarization: Negative for acute ischemic ST-segment deviation or localized T-wave inversion.
\item \textbf{LAO/LAE}: rhythm: Sinus rhythm at a regular rate; morphology: P mitrale with notched P wave in lead II or terminal negative P component in V1 $\geq$ 1 mm deep and $\geq$ 40 ms duration; ST-T repolarization: ST-segments are isoelectric; no pathologic Q waves suggesting old myocardial infarction.
\item \textbf{LQTS}: rhythm: Sinus rhythm with prolonged ventricular repolarization; morphology: Prolonged QTc interval $>$ 470 ms measured in lead II or V5-V6 with normal QRS duration; ST-T repolarization: Absence of acute ST-T abnormalities; T-waves may be broad but lack a specific ischemic territorial pattern.
\item \textbf{LVH}: rhythm: Regular sinus rhythm; morphology: Increased QRS voltage (S V1 + R V5-V6 > 35 mm) with left axis deviation; ST-T repolarization: Asymmetric downsloping ST-segment depression and T-wave inversion in lateral leads I, aVL, V5-V6 representing a left ventricular strain pattern.
\item \textbf{LVOLT}: rhythm: Sinus rhythm with attenuated signal amplitude; morphology: QRS voltage < 5 mm in all limb leads and < 10 mm in all precordial leads; ST-T repolarization: No evidence of territorial ST-elevation or depression; T-waves are concordant but low in amplitude.
\item \textbf{NORM}: rhythm: Normal sinus rhythm 60-100 bpm with consistent P-P intervals; morphology: Normal P wave, PR interval, QRS duration, and QRS axis; ST-T repolarization: No diagnostic ST-segment elevation, depression, or T-wave inversion; no pathologic Q waves in any lead.
\item \textbf{RAO/RAE}: rhythm: Sinus rhythm with prominent atrial signals; morphology: Tall peaked P wave > 2.5 mm in inferior leads II, III, aVF or initial positive P in V1 > 1.5 mm; ST-T repolarization: Negative for acute myocardial injury or primary repolarization abnormalities.
\item \textbf{RVH}: rhythm: Sinus rhythm with rightward QRS vector; morphology: Right axis deviation with dominant R wave in V1 (R/S > 1) and deep S wave in lateral leads V5-V6; ST-T repolarization: ST-segment depression and T-wave inversion in right precordial leads V1-V3 consistent with right ventricular strain.
\item \textbf{STTC}: rhythm: Stable sinus rhythm; morphology: Normal QRS morphology and axis; ST-T repolarization: Nonspecific ST-T abnormalities including minor ST-segment flattening or T-wave inversion without a specific coronary artery territory or reciprocal ST-elevation.
\item \textbf{TAB\_}: rhythm: Sinus rhythm with normal intervals; morphology: QRS duration and P-wave morphology are unremarkable; ST-T repolarization: Generalized T-wave flattening or inversion in contiguous leads without significant ST-segment deviation, excluding acute coronary syndromes or localized injury.
\end{itemize}

\end{document}